\documentclass[pdflatex,sn-mathphys-num]{sn-jnl}% Basic Springer Nature Reference Style/Chemistry Reference Style

\usepackage{svg}
\usepackage{graphicx}%
\usepackage{multirow}%
\usepackage{amsmath,amssymb,amsfonts}%
\usepackage{amsthm}%
\usepackage{mathrsfs}%
\usepackage[title]{appendix}%
\usepackage{xcolor}%
\usepackage{textcomp}%
\usepackage{manyfoot}%
\usepackage{booktabs}%
\usepackage{adjustbox}
\usepackage{algorithm}%
\usepackage{algorithmicx}%
\usepackage{algpseudocode}%
\usepackage{listings}%
\usepackage{tikz}
\usepackage{pifont}
\usepackage{hyperref}
\usepackage{subcaption} % in preamble

\theoremstyle{thmstyleone}%
\theoremstyle{thmstyletwo}%

\theoremstyle{thmstylethree}%

\begin{document}

% \title[Article Title]{Budget-Constrained Active Acquisition for POCUS Echocardiography: Uncertainty-Aware Multi-Task Heart-Failure Assessment}

% \title[Article Title]{Budget-Constrained Scanning for POCUS Echocardiography: Uncertainty-Aware Multi-Task Heart-Failure Assessment}

\title[Article Title]{Optimizing Point-of-Care Ultrasound Video Acquisition for Probabilistic Multi-Task Heart Failure Detection}

%%=============================================================%%
%% GivenName	-> \fnm{Joergen W.}
%% Particle	-> \spfx{van der} -> surname prefix
%% FamilyName	-> \sur{Ploeg}
%% Suffix	-> \sfx{IV}
%% \author*[1,2]{\fnm{Joergen W.} \spfx{van der} \sur{Ploeg} 
%%  \sfx{IV}}\email{iauthor@gmail.com}
%%=============================================================%%

\author*[1]{\fnm{Armin} \sur{Saadat}}\email{arminsdt@ece.ubc.ca}

\author[1]{\fnm{Nima} \sur{Hashemi}}%\email{nima.hashemi.edu@gmail.com}

\author[1]{\fnm{Bahar} \sur{Khodabakhshian}}%\email{baharkhd@ece.ubc.ca}

\author[2]{\fnm{Michael Y.} \sur{Tsang}}%\email{michael.tsang@ubc.ca}

\author[2]{\fnm{Christina} \sur{Luong}}%\email{Christina.Luong@ubc.ca}

\author[2]{\fnm{Teresa S.M.} \sur{Tsang}}%\email{t.tsang@ubc.ca}
\equalcont{These authors jointly supervised the work.}

\author*[1]{\fnm{Purang} \sur{Abolmaesumi}}\email{purang@ece.ubc.ca}
\equalcont{These authors jointly supervised the work.}

\affil[1]{\orgdiv{Department of Electrical and Computer Engineering}, \orgname{The University of British Columbia}, \orgaddress{\city{Vancouver}, \state{BC}, \country{Canada}}}

\affil[2]{\orgname{Vancouver General Hospital}, \orgaddress{\city{Vancouver}, \state{BC}, \country{Canada}}\footnote{T.S.M. Tsang and P. Abolmaesumi are joint senior authors.}}

%%================================%%
%% Sample for structured abstract %%
%%================================%%

\abstract{
\textbf{Purpose:} Echocardiography with point-of-care ultrasound (POCUS) must support clinical decision-making under tight bedside time and operator-effort constraints. We introduce a personalized, budget-constrained data acquisition strategy in which a reinforcement-learning agent, given a partially observed multi-view study, selects the next view to acquire or terminates acquisition to support heart-failure (HF) assessment. 
Upon termination, a diagnostic model jointly predicts aortic stenosis (AS) severity and left ventricular ejection fraction (LVEF), two key HF biomarkers, and outputs calibrated uncertainty, enabling an explicit trade-off between diagnostic performance and acquisition cost.
\textbf{Methods:} We model bedside POCUS as a sequential, cost-constrained acquisition problem: at each step, a video selector (RL agent) chooses the next view to acquire or terminates acquisition. 
Upon termination, a shared multi-view transformer performs multi-task inference with two heads, ordinal AS classification, and LVEF regression, and outputs Gaussian predictive distributions yielding ordinal probabilities over AS classes and EF thresholds. 
These probabilities drive a reward that balances expected diagnostic benefit against acquisition cost (e.g., number of acquired views), producing patient-specific acquisition pathways rather than a fixed protocol.
The video selector is trained with online RL, updating on-policy in a partial-observation simulator built from complete multi-view POCUS studies.
\textbf{Results:} The dataset comprises 12,180 patient-level studies, split into training/validation/test sets (75/15/15). On the 1,820 test studies, our method matches full-study performance while using $32\%$ fewer videos, achieving $77.2\%$ mean balanced accuracy (bACC) across AS severity classification and LVEF estimation, demonstrating robust multi-task performance under acquisition budgets.
\textbf{Conclusion:} Patient-tailored, cost-aware acquisition can streamline POCUS workflows while preserving decision quality, producing interpretable scan pathways suited to bedside use. The framework is extensible to additional cardiac endpoints and merits prospective evaluation for clinical integration. The code is available at \href{https://github.com/Armin-Saadat/Double-Precise}{\texttt{https://github.com/Armin-Saadat/Double-Precise}}}


\maketitle

\section{Introduction}
%%%%%%%%%%%%    Heart Failure   %%%%%%
Heart failure (HF) is a looming epidemic carrying a significant mortality risk, with a 5- and 10-year death rate of 43\% and 65\%, respectively~\cite{jones2019survival}. HF occurs when the heart cannot pump enough blood to meet the body’s needs, resulting in persistent shortness of breath, fatigue, and/or fluid retention that, when severe, may require hospitalization and even cardiac transplantation. Determining the underlying cause of HF is critical, as treatment often diverges substantially; misattribution delays appropriate management and may prolong patient suffering or precipitate premature death. 

%%%%%%%%%%%%    Echocardiography   %%%%%%
Currently, patients must await cardiac imaging with formal echocardiography (echo) to establish the form of HF before targeted treatment can occur. Echo comprehensively evaluates cardiac structure and function, characterizing contractile performance via left ventricular ejection fraction (LVEF) and assessing valve function. Echo findings enable classification of HF as HFrEF (LVEF\(<\)40\%), HFmEF ($\leq40\%$ LVEF $\leq50\%$), HFpEF (LVEF\(>\)50\%), or severe valve disease (aortic or mitral), thereby initiating the treatment. A rapid, accurate, and accessible tool to evaluate the heart for HF, including jointly assessing LVEF and valve disease, could meaningfully transform patient care.

%%%%%%%%%%%%    POCUS   %%%%%%
Access to echo is limited outside urban centers because image acquisition and interpretation require scarce expertise, yielding wait times up to a year~\cite{sanfilippo2005guidelines, munt2006access} and a clear care gap. Point-of-care ultrasound (POCUS) can help bridge this gap by enabling focused scans performed by non-experts on limited-function devices~\cite{luong2019focusus}. However, POCUS users often lack the expertise and time for efficient scanning and precise measurement/interpretation. Machine learning (ML) can streamline acquisition and analysis by guiding operators to capture optimal echo views and HF characterization, thereby facilitating diagnosis while reducing acquisition time. 

%%%%%%%%%%%%    Other AS or EF models   %%%%%%
Prior work on valve disease detection (such as aortic stenosis (AS))~\cite{ahmadi2023transformer, vaseli2023protoasnet, wu2025multiasnet} and LVEF estimation~\cite{mokhtari2024gemtrans, mokhtari2022echognn, behnami2020cine} treats these tasks in isolation and assumes a fixed set of echo cine loops is already available. First, single-task designs ignore inter-task dependencies and physiologic constraints essential for coherent HF assessment. Second, these methods ignore the acquisition stage. Therefore, the POCUS operator must first collect a set of cine loops without guidance on which views to obtain or when to stop, only to get a diagnostic label from models without indicating whether the data are sufficient or what to acquire next. Our prior work, Precise-AS~\cite{saadat2025preciseas} is the only related work that incorporates echo clip acquisition, but optimizes it solely for AS classification and over a limited set of echo views.

%%%%%%%%%%%%    Our proposed method   %%%%%%
% We introduce \textbf{Double-Precise}, a reinforcement learning (RL) framework that optimizes echo acquisition for joint LVEF regression and AS severity classification. It integrates information from multiple cardiac views and yields predictive distributions, enabling uncertainty-aware multitask inference and a coherent joint probability over LVEF and AS. Double-Precise models active video acquisition as a budgeted sequential decision process and uses an RL agent to optimize the balance between diagnostic performance and acquisition time. At each step, the agent estimates the expected diagnostic gain given the already acquired videos and either requests the next video or terminates. The learned acquisition policy is tailored to each patient, which results in different diagnostic pathways across the population.

We introduce \textbf{Double-Precise}, a framework that optimizes echo acquisition for joint LVEF regression and AS severity classification. Starting from no acquired views, a video selector sequentially chooses the next view to acquire or terminates acquisition. Upon termination, a multi-view diagnostic model integrates the acquired views and outputs predictive distributions, enabling uncertainty-aware multitask inference and a coherent joint probability over LVEF and AS.
We train the diagnostic model via supervised learning and keep it frozen. The video selector is then optimized with RL by maximizing a reward that balances diagnostic utility (from the diagnostic model’s outputs) against acquisition cost (e.g., number of acquired videos). To emulate point-of-care scanning retrospectively, we use studies with five available views and model acquisition by starting from a fully masked study and unmasking views as they are selected.

%%%%%%%%%%%%    Summary of results   %%%%%%
Evaluated on a dataset of 12,180 patients, Double-Precise attains high accuracy on both tasks while consistently selecting fewer echo videos across varied acquisition budgets. These results demonstrate the promise of active video acquisition for multitask medical imaging, enabling a fast and reliable initial assessment of HF at the point of care. Our key contributions are as follows.
\begin{itemize}[label=$\bullet$]
\item A transformer-based multi-view multi-task diagnostic model, supporting both classification and regression tasks.
\item An RL framework for active video selection to optimize performance under varying budget constraints. 
\item Confidence-aware diagnosis by modeling the joint probability distribution of AS severity classes and LVEF values.
\end{itemize}

\section{Background}

\subsection{Uncertainty Modelling}
\label{back:uncertainty}
Uncertainty can be categorized into epistemic and aleatoric uncertainties~\cite{kendall2017uncertainties}. Epistemic uncertainty reflects limited knowledge about the model (e.g., due to finite data) and is, in principle, reducible by collecting more informative observations. In contrast, aleatoric uncertainty arises from inherent randomness or measurement noise in the data-generating process and is not reducible by increasing dataset size~\cite{liao2020modelling}. Let the dataset be denoted by $D=\{X,Y\}$, where $X=\{x_i\}_{i=1}^{|D|}$ and $Y=\{y_i\}_{i=1}^{|D|}$ are the sets of $|D|$ observed samples and corresponding labels. The posterior over model parameters $W$ is
$p(W\mid D)=\frac{p(Y\mid X,W)\,p(W\mid X)}{p(Y\mid X)}$.
To model aleatoric uncertainty, we assume that each $y_i$ is a random draw from a distribution $e_i$ associated with the study $x_i$.

Under a Gaussian error model, least-squares regression is equivalent to maximum likelihood estimation~\cite{nix1994estimating}. Specifically,
\begin{equation}
e_i = \mathcal{N}\!\bigl(\mu_i,\sigma_i^2\bigr), 
\quad \mu_i = h(x_i; W_h), 
\quad \sigma_i = g(x_i; W_g) > 0,
\quad W = W_h \cup W_g .
\end{equation}
Here, $h$ and $g$ produce the mean and standard deviation, respectively. Rather than using a single network with parameters $W$ to predict $y_i$ directly, we parameterize the conditional distribution via two models: $h$ (with parameters $W_h$) predicts the mean of $e_i$, and $g$ (with parameters $W_g$) predicts its standard deviation. We then sample
\begin{equation}
y_i \sim \mathcal{N}\!\bigl(h(x_i; W_h),\, g^2(x_i; W_g)\bigr).
\end{equation}

\subsection{Reinforcement Learning}
RL addresses sequential decision making by learning a policy that maximizes long-term reward from interaction with an environment. A standard formalism is the Markov decision process (MDP)~\cite{Littman2001}. 
In MDP, an episode is a trajectory of states, actions, and rewards that begins at an initial state and terminates upon reaching a terminal state. Formally, an MDP is the tuple $(S, S_{end}, A, R, T)$, where $S$ is the state set, $S_{end}\subset S$ the terminal states, and $A$ the action set. The transition function $T: S \times A \times S \rightarrow \mathbb{R}$ specifies the probability of moving to a next state given the current state and action, and the reward function $R: S \times A \times S \rightarrow \mathbb{R}$ assigns the corresponding scalar reward. A policy $\pi: S \rightarrow A$ selects actions from states, and the objective is to identify an optimal policy $\pi^\star$ that maximizes the expected cumulative reward. When the dynamics $T$ and rewards $R$ are unknown but sampleable, RL algorithms estimate $\pi^\star$ from data. Methods are broadly \emph{model-based}, which learn (or use) a model of $T$ and $R$ to plan, and \emph{model-free}, which optimize value functions and/or policies directly from experience. Canonical model-free approaches include value-based methods (e.g., Q-learning, which seeks $Q^\star$) and policy-gradient/actor–critic methods, which optimize a parameterized policy $\pi_\theta$ by estimating $\nabla_\theta J(\pi_\theta)$. Deep RL augments these families with function approximation to handle high-dimensional state and action spaces (e.g., deep Q-networks~\cite{mnih2013atari}; proximal policy optimization~\cite{schulman2017ppo}).
\section{Method}

\begin{figure}[t]
    \centering
    \includegraphics[width=1\linewidth]{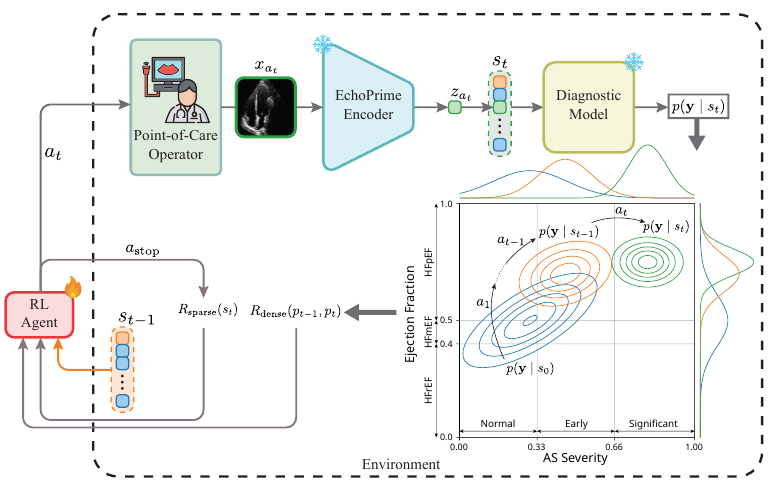}
    \caption{High-level overview of Double-Precise as an RL framework. The agent sequentially selects videos until a stop action; the environment, comprising a video encoder and a diagnostic model, returns sparse and dense rewards and next states. The diagnostic model outputs a joint distribution over AS severity and LVEF categories; the blue contour depicts this distribution at the initial state, which shifts as additional videos are acquired. As acquisition proceeds, the marginal variances contract, indicating increasing diagnostic certainty.
}
    \label{fig:architecture}
\end{figure}

Following prior work on active feature acquisition~\cite{li2021towards,yin2020reinforcement}, we cast video acquisition as a discrete-time RL problem (Figure~\ref {fig:architecture}).
In \emph{Double-Precise}, an RL agent, hereafter the \emph{video selector}, interacts with an environment composed of two components: a \emph{video encoder} and a \emph{diagnostic model}. The process begins from a zero-masked feature set ($s_0$). At step $t$, the video selector takes action $a_t \in \{1, N\}$, where $N$ is the maximum number of videos per patient. The video encoder then takes in the selected video $x_{a_t} \in \mathbb{R}^{H \times W \times 3}$ and outputs its embedding $f(x_{a_t}) \in \mathbb{R}^{D}$, where $D$ is the embedding size. This takes us to state $s_t \in \mathbb{R}^{N \times D}$: 
\begin{equation}
\label{eq:state}
\mathrm{s}_{t} = [z_1, \dots, z_N], 
\quad\mathrm{where}\quad
z_i = \bigl[m_i \cdot f(x_i)\bigr], 
\quad
m_i =
\begin{cases}
1, & \text{if } i \in [a_1, \dots, a_t],\\
0, & \text{otherwise.}
\end{cases}
\end{equation}
The diagnostic model receives $s_t$ and outputs a probabilistic diagnosis; this output defines the reward signal that guides optimization of the data acquisition policy. The new state $s_t$ is subsequently fed back to the video selector, which is used as input for taking the next action $a_{t+1}$. The video selector may terminate acquisition by choosing a dedicated stop action $a_{stop}$. Upon termination, the diagnostic model produces a final prediction from the latest state, i.e., from the set of embeddings corresponding to the acquired videos. The following sections describe each component in detail.

\subsection{Video Encoder}
Double-Precise employs the video encoder from EchoPrime, a multi-video, view-informed vision–language foundation model for echo trained contrastively on over 12 million video–report pairs~\cite{vukadinovic2024echoprime}. EchoPrime comprises a video encoder (mViT pretrained on Kinetics with a 512-dimensional output) trained jointly with a text encoder to learn a unified video–text embedding space; a view classifier (58 standard echo views) and an anatomical attention module that weights clips across views for study-level interpretation. EchoPrime embeddings place all clips from a study in a common latent space, facilitating cross-view comparison and integration. We extract fixed video-level embeddings by passing each clip through the frozen EchoPrime encoder. These embeddings form the RL state representation ($s$) supplied to both the video selector and the diagnostic model.

\subsection{Diagnostic Model}
The model comprises a task-aware encoder and a multivariate Gaussian head. 

\subsubsection{Transformer Encoder}
We adopt a Transformer encoder~\cite{vaswani2017attention} tailored to the input structure $s_t$, a sequence of video embeddings in which missing views are zero-masked. Each video embedding is treated as a token. The encoder applies an attention mask so that masked (missing) videos neither attend to nor receive attention from other tokens, compelling the model to rely on the acquired videos~\cite{saadat2025preciseas}.  To support multi-task study-level prediction (AS severity and LVEF), we append two task tokens to the sequence, following the tokenization strategy of ViT~\cite{dosovitskiy2021vit}. The full token sequence (video tokens + task tokens) is processed by the encoder (\ref{eq:task-tokens}). The task tokens act as study-level aggregators for their respective targets and are passed to the multivariate head.
\begin{equation}
\label{eq:task-tokens}
\textit{token}_{\small{AS}},\;\textit{token}_{\small{EF}} = \textit{TransformerEncoder}(s_t).
\end{equation}
We choose one token per task rather than a single token or mean pooling, because dedicated tokens facilitate optimization of task-specific objectives and improve representational disentanglement~\cite{darcet2024registers}.

\subsubsection{Multivariate Head}
We implement an uncertainty-aware multi-task head that models a joint (multivariate) Gaussian over AS severity and LVEF. Rather than outputting point estimates, the head predicts
\begin{equation}
p(\mathbf{y}\mid s_t)=\mathcal{N}\!\bigl(\boldsymbol{\mu}_{s_t},\,\boldsymbol{\Sigma}_{s_t}\bigr), 
\quad\mathbf{y}=\bigl(y_{\mathrm{\small{AS}}},\,y_{\mathrm{\small{EF}}}\bigr)^{\!\top},
\end{equation}  

\begin{equation} 
\quad \boldsymbol{\mu}_{s_t} = h(\textit{token}_{\small{AS}}\oplus\textit{token}_{\small{EF}}; W_h), 
\quad \boldsymbol{\Sigma}_{s_t} = g(\textit{token}_{\small{AS}}\oplus\textit{token}_{\small{EF}}; W_g),
\end{equation}
where $\boldsymbol{\mu}_{s_t}\in\mathbb{R}^2$ is the mean, $\boldsymbol{\Sigma}_{s_t}\in\mathbb{R}^{2\times 2}$ is the non-diagonal covariance which captures dependence between AS and LVEF, and $\oplus$ is the concatenation operator. $h$ and $g$ are deep logistic regression models and ensure a value range $[0, 1]^2$. 

\subsubsection{Training}
For AS severity classification, we use the marginal over the AS dimension and discretize $[0,1]$ uniformly into $K$ classes. Class probabilities $\hat{\mathbf{p}}\in\mathbb{R}^K$ are obtained by sampling from the AS marginal. Given the ground-truth class label $y_{\mathrm{AS}}\in\{1,\dots,K\}$, we optimize the negative log-likelihood:
\begin{equation}
\mathcal{L}_{\mathrm{AS}}
=
\mathrm{nll}\!\big(y_{\mathrm{AS}},\,\log\hat{\mathbf{p}}\big).
\end{equation}

For LVEF (a regression target), given the ground-truth value $y_{\mathrm{EF}}\in[0, 1]$, we minimize the Gaussian negative log-likelihood:
\begin{equation}
\mathcal{L}_{\mathrm{EF}}
= -\log \mathcal{N}\!\bigl(y_{\mathrm{EF}}\mid \mu_{\mathrm{EF}}, \sigma^2_{\mathrm{EF}}\bigr)
= \tfrac{1}{2}\!\left[\log\!\bigl(2\pi\sigma^2_{\mathrm{EF}}\bigr)
+ \frac{\bigl(y_{\mathrm{EF}}-\mu_{\mathrm{EF}}\bigr)^2}{\sigma^2_{\mathrm{EF}}}\right],
\end{equation}
where $\mu_{\mathrm{EF}}$ and $\sigma^2_{\mathrm{EF}}$ are the EF marginal mean and variance from the joint distribution. Given nonnegative weights $\lambda_{\mathrm{AS}},\lambda_{\mathrm{EF}}$, the total loss is: 
\[
\mathcal{L}_{\text{total}}
= \lambda_{\mathrm{AS}}\,\mathcal{L}_{\mathrm{AS}}
+ \lambda_{\mathrm{EF}}\,\mathcal{L}_{\mathrm{EF}}.
\]

We train the diagnostic model in an end-to-end manner using supervised learning on full-acquisition, study-level inputs. 
After convergence, the diagnostic model is frozen, and the predicted joint probabilities are used as a dense reward for optimizing the acquisition policy. 

\subsection{Video Selector}

\subsubsection{Network}
Double-Precise optimizes the video acquisition policy via reinforcement learning, following prior work~\cite{gabriel2022rl,muyama2024deep,yu2023deep,saadat2025preciseas,kim2025learningstop}. 
We adopt proximal policy optimization (PPO)~\cite{schulman2017ppo} with an \emph{actor} (policy) and a \emph{critic} (value function). 
Given the state $s_t \in \mathbb{R}^{N\times D}$, the actor $f_\theta:\mathbb{R}^{N\times D}\!\to\!\mathbb{R}^{|\mathcal{A}|}$ produces action logits, inducing a stochastic policy $\pi_\theta(a\mid s_t)$ from which $a_{t+1}$ is sampled. 
The critic $g_\phi:\mathbb{R}^{N\times D}\!\to\!\mathbb{R}$ estimates the state value. 
Both actor and critic are implemented as multi-layer perceptrons; the actor ends with a linear logits head over actions, and the critic with a scalar regression head. Each action $a_i$ selects a specific video, and $a_{\textit{stop}}$ terminates the acquisition, therefore; $|\mathcal{A}| = N+1$. 

\subsubsection{Reward}
We incorporate both sparse and dense reward schemes. 
At each step, we assign a dense reward based on the change in the diagnostic model’s joint predictive distribution. 
Let $p_{t} = p(\mathbf{y}\mid s_t)$ denote the joint PMF over (AS, LVEF) at step $t$; the reward is the Jensen–Shannon (JS) divergence between successive predictions:
\begin{equation}
R_{\textit{dense}} = \mathrm{JS}\!\bigl(p_{t-1}\,\|\,p_{t}\bigr),
\;
\mathrm{JS}(p\|q) = \tfrac{1}{2}\mathrm{KL}\!\bigl(p\|m\bigr) + \tfrac{1}{2}\mathrm{KL}\!\bigl(q\|m\bigr),
\; m=\tfrac{1}{2}(p+q),
\end{equation}
where KL is the Kullback–Leibler divergence. This dense reward encourages acquisitions that yield informative distributional updates. Upon termination by selecting $a_{\textit{stop}}$ at state $s_t$, we assign a sparse terminal reward
\begin{equation}
\label{eq:sparse-reward}
R_{\textit{sparse}}(s_t)
= \mathbf{1}\!\bigl\{\,y_{\small{AS}}^{\mathrm{pred}}(s_t)=y_{\small{AS}}^{\mathrm{true}}\,\bigr\}
+
\mathbf{1}\!\bigl\{\,y_{\small{EF}}^{\mathrm{pred}}(s_t)=y_{\small{EF}}^{\mathrm{true}}\,\bigr\}
\;-\; \lambda \sum_{i=1}^{t} c_{a_i},
\end{equation}
where $\mathbf{1}\{\cdot\}$ is the indicator of correct diagnosis, $c_{a_i}$ is the cost of action $a_i$, and $\lambda>0$ is a cost coefficient. 
Here, we prioritize the clinical category of LVEF rather than its exact value; $y_{\mathrm{EF}}$ denotes the corresponding category.
The video selector may take at most $|\mathcal{A}|$ actions; if it exhausts all $|\mathcal{A}|$ actions without invoking $a_{\textit{stop}}$, the episode terminates with zero sparse reward.

\section{Experiments and Results}
\subsection{Dataset}
We conducted experiments on a private echo dataset drawn from a tertiary-care hospital’s study archive under institutional review board approval. Videos were acquired on Philips iE33 and GE Vivid i/E9 systems. A view-detection algorithm~\cite{liao2019modelling} automatically identified parasternal long-axis (PLAX), parasternal short-axis (PSAX), and apical two-, three-, and four-chamber views (AP2, AP3, AP4). The dataset comprises 12,180 patient-level studies and was partitioned into training, validation, and test sets in a 75:15:15 ratio with patient-disjoint splits. AS labels derived from Doppler measurements yielded 4,503 normal, 3,173 early, and 4,504 significant cases. Based on LVEF, studies were categorized as HFrEF (LVEF $<40\%$, $n=3,552$), HFmEF ($40\%\le\text{LVEF}\le50\%$, $n=4,124$), and HFpEF (LVEF $>50\%$, $n=4,503$). For consistency, we randomly selected one clip per view per study (N=5), and imposed a fixed ordering $[\text{AP2},\text{AP3},\text{AP4},\text{PLAX},\text{PSAX-Ao}]$, ensuring a stable view-to-token correspondence. Each clip was assigned a unit acquisition cost, reflecting the homogeneous imaging modality, so the total study cost is proportional to the number of acquired videos.

\subsection{Implementation Details}
Actor and critic are separate 3-layer MLPs; the policy is trained with PPO for 50 epochs with a discount factor of $1.0$, and the best checkpoint is chosen by balanced accuracy (bACC) on the validation set. The diagnostic Transformer has 3 encoder layers, 4 heads, 32-dim tokens, and a 128-dim feed-forward; it is trained for 50 epochs.
All models are implemented in PyTorch and trained on an NVIDIA B200 GPU.

\subsection{Evaluations}

\begin{table}[h!]
\caption{Quantitative metrics include balanced accuracy (bACC)\%, weighted F1\%, and balanced mean absolute error (bMAE) for the regression task. The results are study-level, with the best results in bold.}
\label{tab:metric}
\centering
\begin{tabular*}{\textwidth}{@{\extracolsep\fill}c|c|cccccc}
\toprule
\multirow{2}{*}{Acquisition} & \multirow{2}{*}{Selection Method} & \multicolumn{2}{@{}c@{}}{Aortic Stenosis} & \multicolumn{3}{@{}c@{}}{LV Ejection Fraction} & \multirow{2}{*}{Mean}\\
\cmidrule(lr){3-4}\cmidrule(lr){5-7}
Ratio &  & bACC$\uparrow$ & F1$\uparrow$ & bACC$\uparrow$ & F1$\uparrow$ & bMAE$\downarrow$ & bACC\\ 
\midrule
%%%%%%%%%%%%%%%%%%%%%%%%%%%%%
\multirow{3}{*}{$1/5$} & Random & 53.8 & 53.9 & 57.1 & 57.3 & 9.13 & 55.4\\
 & Pop-wise & 68.6 & 70.4 & 69.3 & 69.8 & 7.20 & 69.0 \\
 & RL & \textbf{69.3} & \textbf{71.1} & \textbf{69.8} & \textbf{70.4} & \textbf{6.97} & \textbf{69.6}\\
 \midrule
 %%%%%%%%%%%%%%%%%%%%%%%%%%%%%
 \multirow{3}{*}{$2/5$} & Random & 65.6 & 67.3 & 69.7 & 70.5 & 7.03 & 67.6 \\
 & Pop-wise & 72.4 & 74.4 & 72.2 & 73.1 & 6.52 & 72.3\\
 & RL & \textbf{73.8} & \textbf{75.7} & \textbf{73.1} & \textbf{73.8} & \textbf{6.38} & \textbf{73.4}\\
 \midrule
 %%%%%%%%%%%%%%%%%%%%%%%%%%%%%
 \multirow{3}{*}{$3/5$} & Random & 72.1 & 73.9 & 74.8 & 75.5 & 6.04 & 73.4\\
 & Pop-wise & 73.8 & 75.8 & 77.1 & 77.5 & \textbf{5.59} & 75.5 \\
 & RL & \textbf{75.9} & \textbf{77.7} & \textbf{77.6} & \textbf{78.1} & 5.81 & \textbf{76.7}\\
 \midrule
 %%%%%%%%%%%%%%%%%%%%%%%%%%%%%
 \multirow{3}{*}{$4/5$} & Random & 74.8 & 76.7 & 77.5 & 78 & 5.54 & 76.1\\
 & Pop-wise & 76.6 & 78.5 & 76.4 & 76.8 & 5.55 & 76.5\\
 & RL & \textbf{75.9} & \textbf{77.8} & \textbf{78.4} & \textbf{79} & \textbf{5.43} & \textbf{78.4}\\
\bottomrule
\end{tabular*}
\end{table}

\begin{figure}[t]
  \centering
  \begin{subfigure}{0.5\linewidth}
    \centering
    \includegraphics[width=\linewidth]{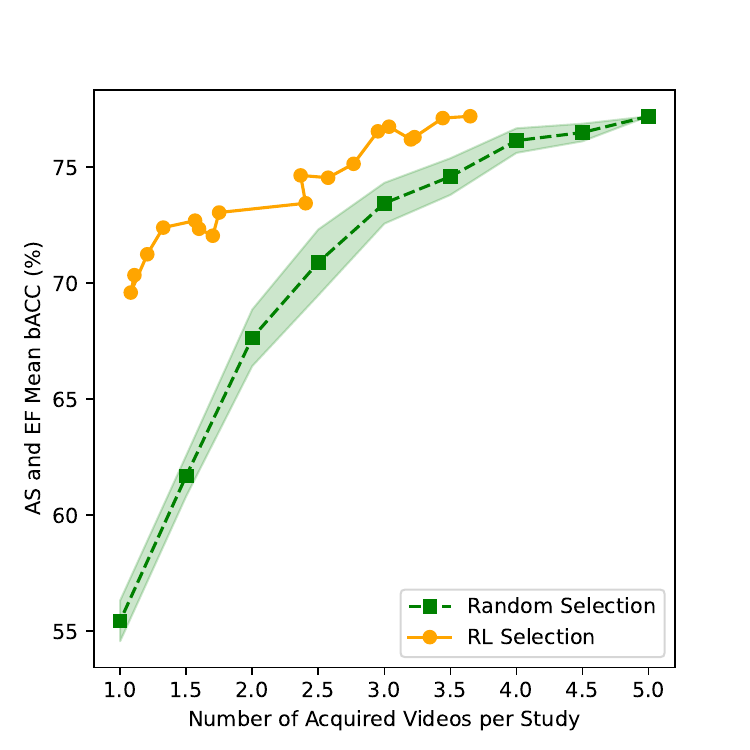}    
    \label{fig:meanbacc}
  \end{subfigure}\hfill
  \begin{subfigure}{0.5\linewidth}
    \centering
    \includegraphics[width=\linewidth]{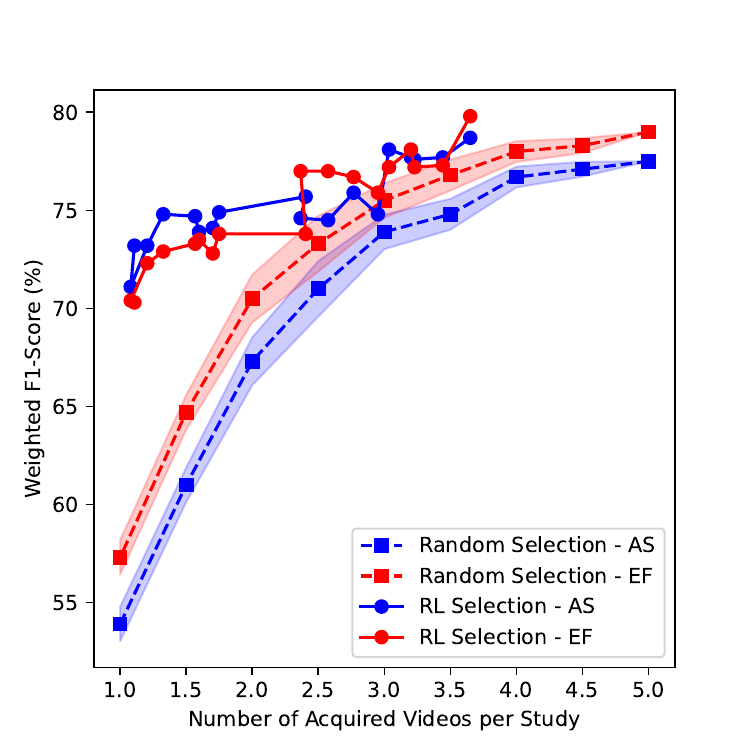}
    \label{fig:f1}
  \end{subfigure}
  \caption{Comparison of RL and random selection. Each circle point corresponds to a $\lambda$ from Double-Precise. For random selection, the dashed lines represent the mean performance over 5 runs, while the shaded regions indicate one standard deviation.}
  \label{fig:metric}
\end{figure}

Table~\ref{tab:metric} summarizes the performance of \textit{Double-Precise} versus two baselines under varying acquisition budgets. 
\emph{Random} selects \(k\) videos at random per study. 
\emph{Pop-wise} applies the same subset of \(k\) views, the combination that achieves the best validation performance, to all patients (i.e., population-level, non-patient-specific selection).
Across budgets and metrics, RL-based selection outperforms both baselines.

Fig.~\ref{fig:metric} compares random versus RL-based video selection. RL consistently achieves a superior performance–efficiency trade-off, attaining the best classifier performance with fewer inputs: random selection requires approximately five videos per study, whereas RL achieves comparable or better results with about 3.4 videos.

\begin{figure}[t]
    \centering
    \includegraphics[width=1\linewidth]{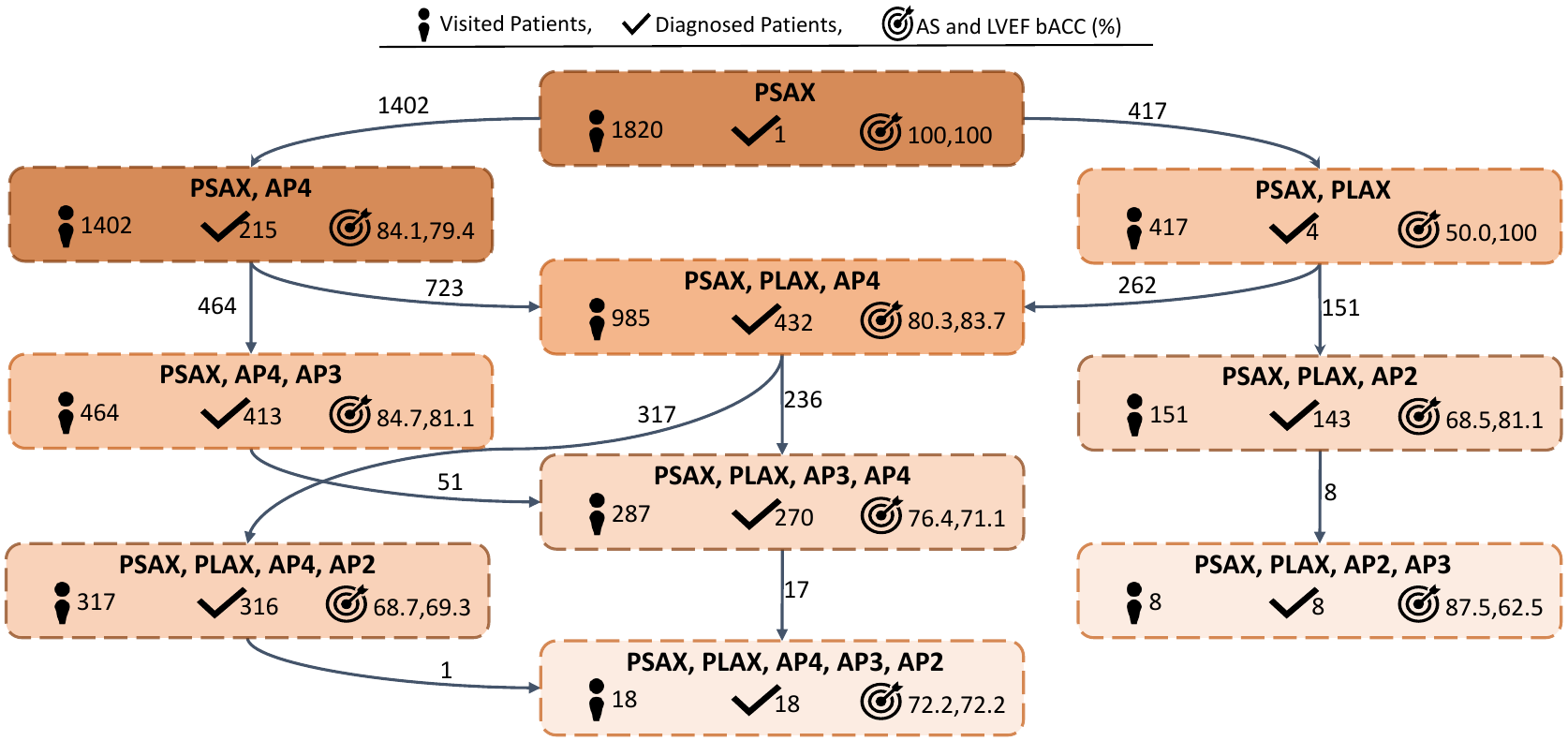}
    \caption{Personalized diagnostic pathways for joint AS and LVEF prediction. Each node corresponds to a state defined by the set of acquired videos, and reports the number of patients reaching that state, the number whose acquisition terminates there, and the bACC for AS and LVEF among patients whose acquisition terminates at that node. Directed edges represent the sequential view acquisitions between states, annotated by the number of patients transitioning along each edge.}
    \label{fig:pathways}
\end{figure}

Fig.~\ref{fig:pathways} visualizes the personalized acquisition pathways, showing how patients transition between partial-view states and where the policy chooses to terminate. By aggregating the number of patients along each branch and the terminal bACC at each stopping state, the figure highlights which view sequences the model relies on most and how diagnostic performance varies across pathways. This provides an interpretable summary of the sequential decision process and can help identify high-yield views under constrained acquisition budgets.

\subsection{Ablation Study}
\begin{table}[t]
\caption{
Ablation of the cost coefficient $\lambda$, illustrating the trade-off between acquisition efficiency and diagnostic performance (larger $\lambda$ favors fewer acquisitions). Reported $\mathrm{bACC}$ and $\mathrm{F1}$ are averaged over AS and LVEF. Each cell contains the mean and std over 5 runs. Entries are mean~$\pm$~std over 5 runs.
}
\label{tab:ablation}
\centering
\begin{tabular}{c|ccc|cc}
\multirow{2}{*}{\shortstack{$\lambda$}}
& \multicolumn{3}{c}{Study-level}\vline
& \multicolumn{2}{c}{Acquired-Videos }
\\ 
& \multicolumn{1}{c}{~~bACC$\uparrow$~~} 
& \multicolumn{1}{c}{~~~~~F1$\uparrow$~~~~~} 
& \multicolumn{1}{c}{~~Reward$\uparrow$~~}\vline
& \multicolumn{1}{c}{~~Ratio$\downarrow$} 
& \multicolumn{1}{c}{~~Count$\downarrow$}
\\ 
\hline 
\hline
w/o RL
& 77.2(0.0)
& 78.5(0.0)
& N/A
& 100\%  
& 5.00
\\
0.001
& \textbf{77.2(0.3)}
& \textbf{78.6(.01)}
& \textbf{1.54(.1)}
& 68\%  
& 3.40
\\
0.01
& 75.1(0.4)
& 76.4(.02)
& 1.44(.2)
& 55\% 
& 2.76
\\
0.05
& 73.0(0.3)
& 74.2(.01)
& 1.31(.3)
& 30\%  
& 1.75
\\
0.1
& 72.4(0.6)
& 74.0(.03)
& 1.21(.3)
& 24\% 
& 1.20
\\
0.2
& 70.1(0.5)
& 71.6(.03)
& 1.01(.02)
& 21\%
& 1.08
\\
0.5
& 33.3(0.1)
& 33.3(.00)
& 0.05(.01)
& 0\%
& 0.01
\end{tabular}
\end{table}

The cost coefficient ($\lambda$) in Eq.~\ref{eq:sparse-reward} plays a crucial role in training the RL agent. Increasing $\lambda$ causes the model to select fewer videos, which in turn reduces performance (see Table~\ref{tab:ablation}). If $\lambda$ is set excessively high, the RL agent terminates the acquisition process immediately without selecting any videos.

The cost coefficient $\lambda$ in Eq.~\ref{eq:sparse-reward} governs the efficiency–accuracy trade-off during RL training. Larger $\lambda$ penalizes acquisitions more heavily, leading the agent to select fewer videos and, consequently, to reduced diagnostic performance (Table~\ref{tab:ablation}). In the extreme, an excessively large $\lambda$ drives immediate termination without selecting any video.

\section{Conclusion}
Double\mbox{-}Precise casts bedside echocardiography as active video acquisition, coupling a study-level diagnostic model with an RL policy to balance diagnostic gain against acquisition cost. Leveraging EchoPrime embeddings, a task-aware Transformer, and a joint probabilistic head over AS and LVEF, it produces calibrated, study-level predictions while adaptively selecting patient-specific views. Empirically, it matches fixed-protocol performance with fewer videos ($\approx 32\%$ reduction), indicating that policy-driven acquisition can streamline POCUS without sacrificing decision quality, although evaluations are retrospective on pre-acquired studies, so the policy is not tested in a live bedside POCUS workflow. Under stringent acquisition budgets, where accuracy inevitably falls below the full-acquisition upper bound, the RL policy preserves the most performance relative to baselines. By prioritizing the most informative patient-specific views, it achieves the highest retained performance for a given budget and degrades most slowly as the budget tightens, an important property for bedside POCUS, where efficiency is often required under operator variability. Future work on prospective POCUS studies will help confirm feasibility and deployability.\\

%%%%%%%%%%%%%%%%%%%% Acknowledgments

\noindent{\textbf{Acknowledgments.} This research was supported in part by the Canadian Institutes of Health Research (CIHR) and the Natural Sciences and Engineering Research Council of Canada (NSERC), and through computational resources and services provided by Advanced Research Computing at the University of British Columbia.\\}

\noindent\textbf{Disclosure of Interests.}
The authors have no competing interests to declare.

\bibliography{sn-bibliography}

\end{document}